\documentclass{article} 
\usepackage{iclr2016_conference,times}
\usepackage{hyperref}
\usepackage{latexsym}
\usepackage{epsfig}
\usepackage{graphicx}
\usepackage{amsmath}
\usepackage{amssymb}
\usepackage{cool}
\usepackage{enumitem}

\title{Multimodal Skip-gram Using Convolutional Pseudowords}

\author{Zachary Seymour\\
Binghamton University, SUNY\\
{\tt zseymou1@binghamton.edu}\\\And
  Yingming Li\\
Zhejiang University\\
{\tt liymn@zju.edu.cn}\\\And
  Zhongfei (Mark) Zhang \\
  Binghamton University, SUNY\\
  {\tt zhongfei@cs.binghamton.edu}
}

\date{}

\begin{document}
\maketitle

\begin{abstract}
   This work studies the representational mapping across multimodal data
such that given a piece of the raw data in one modality  the
corresponding semantic description in terms of the raw data in another
modality is immediately obtained.  Such a representational mapping can be found in
a wide spectrum of real-world applications including image/video
retrieval, object recognition, action/behavior recognition, and event
understanding and prediction. To that end, we introduce a simplified
training objective for learning multimodal embeddings using the
skip-gram architecture by introducing convolutional ``pseudowords:''
embeddings composed of the additive combination of distributed word
representations and image features from convolutional neural networks
projected into the multimodal space. We present extensive
results of the representational properties of these embeddings on
various word similarity benchmarks to show the promise of this approach.
\end{abstract}

\section{Introduction}
Distributed representations of multimodal
embeddings~\citep{feng2010visual} are receiving
increasing attention recently in the machine learning literature, and
techniques developed have found a wide spectrum of applications in the
real world.
These types of vector representations are particularly desirable for
the way in which they better model the grounding of perceptual or semantic
concepts in human vocabulary~\citep{lazaridou2015combining,
  glenberg2000symbol,hill2014learning}.

As such, there has been development towards so-called \emph{multimodal
distributional semantic models}~\citep{silberer2014learning,
lazaridou2015combining, kiros2014multimodal, frome2013devise,
bruni2014multimodal}, which leverage textual
co-occurance and visual features to form multimodal representations of
words or concepts.

The work introduced in~\citet{lazaridou2015combining} sought to
address many of the drawbacks of these models.
In particular, by incorporating visual information into the training
objective, they address the biological inaccuracy of the existing models,
in that word representations grounded in visual information
have been shown to more closely approximate the way humans learn language.
Furthermore, incorporating visual information alongside the text
corpus allows the training set to consist of both visual and
non-visual words.
As a result, the induced multimodal representations and multimodal
mapping no longer rely on the assumption of full visual coverage of the
vocabulary, so the results are able to generalize beyond the initial
training set and to be applied to various representation-related tasks, such
as image annotation or retrieval.

In this work, we introduce a further refinement on the multimodal
skip-gram architecture, building upon the approaches of~\citet{mikolov2013efficient,mikolov2013distributed},
\citet{hill2014learning}, and \citet{lazaridou2015combining}. 
Rather than adding a visual term to the linguistic training objective, we
directly situate terms in a visual context by replacing relevant words
with \emph{multimodal pseudowords}, derived by composing the
textual representations with convolutional features projected into the
multimodal space.
In this way, we further address the grounding problem
of~\citet{glenberg2000symbol} by incorporating the word-level
visual modality directly into the sentence context.
This model represents an advancement of the existing literature
surrounding multimodal skip-gram, as well as multimodal distributional
semantic models in general, by greatly simplifying the method of
situating the words in the visual context and reducing the
number of hyperparameters to tune by directly incorporating multimodal
words into the existing objective function and hiearchical softmax
formulations of the skip-gram models.

Finally, we would also like the learned embeddings to be applicable to the problem of \emph{zero-shot
  learning}~\citep{socher2013zero, lazaridou2014wampimuk,
  frome2013devise}.
By incorporating perceptual information into the skip-gram learning
objective, we can leverage vocabulary terms for which no
manually-annotated images were originally available.
In this way, these learned representations can be used to both grow
the annotation set and retrieve new annotations for a given image set.

\section{Related Work}
In the last few years, there has been a wealth of literature
on multimodal representational models.
As explained in~\citet{lazaridou2015combining}, the majority of
this literature focuses on constructing textual and visual representations
independently and then combining them under some metrics.
\citet{bruni2014multimodal}
utilize a direct approach to ``mixing'' the vector representations by
concatenating the text and image vectors and applying Singular Value
Decomposition.
The image vectors used here, though, are constructed using the
bag-of-visual-words method.

In~\citet{kiela2014learning}, the authors utilize a more
sophisticated approach to the concatenation method by extracting
visual features using state-of-the-art convolutional neural networks
and the skip-gram architecture for the text.
Similarly,~\citet{frome2013devise} also utilizes the skip-gram
architecture  and convolutional features; however the two modalities
are then combined using a natural similarity metric.

Other recent work has presented several methods for directly
incorporating visual context in neural language models.
In~\citet{xuimproving}, word context is enhanced by
\emph{global} visual context; \textit{i.e.}, a single image is used
as the context for
the whole sentence (conversely, the sentence acts as a caption for the image).
The multimodal skip-gram architecture proposed
by~\citet{lazaridou2015combining} takes a more fine-grained
approach by incorporating word-level visual context and concurrently
training words to predict other text words in the window as well as their
visual representation.
Our model makes this approach even more explicit, by training the word
vectors to predict an additive composition of the textual and visual
context and thus constructing an implicit mapping between the textual
and visual modalities.

Finally, the work introduced in~\citet{hill2014learning}
employs a similar ``pseudoword'' architecture to that proposed here.
However, the visual features used are in the form of perceptual
information derived from either user-generated attributes or other textual annotations of imagery.
While this is shown to be useful for distinguishing classes of words
(\textit{e.g.}, between abstract and concrete), it precludes any incorporation of
visual, non-linguistic context and thus the derivation of any
mapping between images and words or applications to representation-related tasks.

\section{Architecture}

\subsection{Skip-gram for Word Representations}
\label{sec:skip-gram}
This model is primarily derived from the skip-gram model introduced
by~\citet{mikolov2013linguistic}.
Skip-gram learns representations of words that predict a target word's
context.
The model maximizes
\begin{equation}
  \label{eq:1}
  \frac{1}{T} \Sum{\left(\Sum{\log p(w_{t+j} | w_t)}{{-c\leq j \leq
          c,j \not= 0}} \right)}{t,1,T}
\end{equation}
where \(w_1, w_2, \dotsc, w_T\) are words in the training set and
\(c\) the window size around the target word.
The probablity \(p(w_{t+j} | w_t)\) is given by softmax, that is:
\begin{equation}
  \label{eq:2}
  p(w_{t+j} | w_t) = \frac{e^{{u'_{w_{t+j}}}^T
      u_{w_t}}}{\Sum{e^{{u'_{w'}}^T u_{w_t}}}{w',1,W}}
\end{equation}
where \(u_w\) and \(u'_w\) are the context vector and target vector
representations induced for word \(w\), respectively, and \(W\) gives
the vocabulary size.
To speed up this computation, a Huffman tree is constructed from the
vocabulary, and then the softmax formula is replaced with the hierarchical
softmax~\citep{morin2005hierarchical}.

\subsection{Skip-gram with Multimodal Pseudowords}
\label{sec:skip-gram-multimodal}

To ground the word representations in a visual context, we introduce a
means of replacing a word in the corpus with its corresponding
multimodal pseudoword.
The pseudoword vector for a given word \(w\), denoted \(z_w\), is given by
\begin{equation}
  \label{eq:3}
  z_w = u_w + Mv_w
\end{equation}
where \(u_w\) is the multimodal word representation of \(w\) to be
induced, \(v_w\) is the visual data for the concept represented by
\(w\), and \(M\) is the mapping induced between the visual and
multimodal space. (The sources of the textual and visual data are
explained below.)
Where no visual features are available for a given word, \(v_w\) is set to
0.
Thus, the objective function in~\eqref{eq:1} remains the
same, while each word vector in the context window of the current word
in~\eqref{eq:2} is replaced
with its corresponding pseudoword. 
In this way, each target word in the corpus is trained to predict
every given \emph{pseudoword} in its context window.

\begin{figure}
\centering
\includegraphics[width=0.8\textwidth]{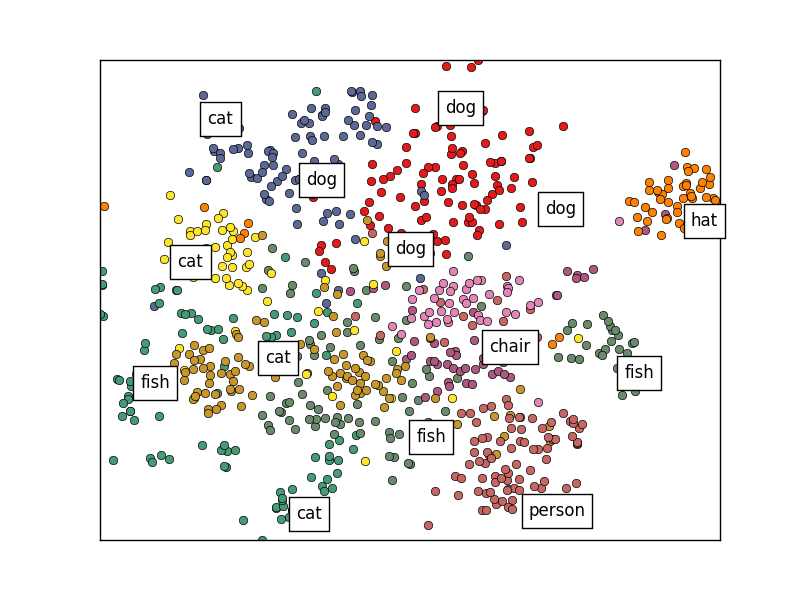}
\caption{t-SNE embedding of a subset of the convolutional features extracted from Imagenet, demonstrating the inherent clustering of the dataset.}
\label{fig:tsne}
\end{figure}

For the value of \(v_w\), a key issue in this approach is selecting a canonical visual representation for a given concept (\textit{e.g.}, a single image labeled ``dog'' does not necessarily accurately represent the visual information of \emph{all} dogs).
Consequently, although the features extracted for a given image from a convolutional neural network (CNN) provided a higher-level visual representation than the raw pixel data, these features are not representative of the concept as a whole.
Rather than handpicking an appropriate image for each class, we rely on the manner in which the CNN features form clusters based on their corresponding visual concepts, a well-explored phenomenon.\footnote{See, for example, \url{http://cs.stanford.edu/people/karpathy/cnnembed/}}
Thus, for each visual word, we can sample some images corresponding to the concept and extract CNN features for the images.
Figure~\ref{fig:tsne} shows the nature of some of these clusters; based on this intuition,
we test two approaches to this problem, which we refer to as the \emph{centroid} method and the \emph{hypersphere} method.

For the \textit{centroid} method\footnote{This is the approach taken by~\citet{lazaridou2015combining}}, the CNN features sampled for a given visual concept are averaged together.
In this way, we form unified representation for each cluster by extracting its centroid and using this for \(v_w\).
We not, however, that this somewhat limits the representational quality of \(v_w\) by condensing the varied class of images to a single data pointl.

To this end, we introduce the \textit{hypersphere} method as a means of capturing the complexity of the visual space.
Rather than averaging the sampled CNN features, we instead fit a Gaussian mixture model to the cluster of CNN features, and, at each training step, sample a point from the model.
This method, then, retains more of the variation between samples present in the dataset, while still only drawing on a small set of original images.
The hypersphere technique can also be seen as a means of augmenting the training data without directly extracting more convolutional features.
Since all points in a given cluster are assumed to be similar and to represent the same concept, a ``new'' image can be used to form \(v_w\) at each training step without necessitating an equivalently-large image dataset.

\section{Experiments}
\subsection{Experimental Data}
\label{sec:experimental-data}
For our text corpus, keeping with the existing literature, we use a preprocessed dump of Wikipedia\footnote{\url{http://wacky.sslmit.unibo.it}} containing approximately 800M tokens.
For the visual data, we use the image data from ILSVRC
2012~\citep{ILSVRC15} and the corresponding Wordnet
hierarchy~\citep{miller1995wordnet} to represent a word
visually if the word or any of its hyponyms has an entry in Imagenet and occurs more than 500 times in the text corpus.
This yields approximately 5,100 ``visual'' words.

To construct the vectors for the visual representations, we follow a similar experimental set-up
as that used by~\citet{lazaridou2015combining}. In each of the cases described above---\textit{centroid} and \textit{hypersphere}---, we randomly sample 100 images from the corresponding synsets of Imagenet for each visual word and use a
pre-trained convolutional neural network as described
in~\citet{krizhevsky2012imagenet} via the Caffe
toolkit~\citep{jia2014caffe} to extract a 4096-dimensional
vector representation of each image.
We then treat the 100 vectors corresponding to each of the 5,100 visual words as clusters in the 4096-dimensional visual space.

\subsection{Approximating Human Similarity Judgments}
\label{sec:appr-human-simil}

\begin{description}[style=unboxed,leftmargin=0cm]
\item[Word Similarity Benchmarks] To compare our technique to the existing literature, we evaluate our
embeddings on four common benchmarks which capture several diverse
aspects of word meaning:
\textit{MEN}~\citep{bruni2014multimodal},
\textit{Simlex-999}~\citep{hill2014simlex},
\textit{SemSim}~\citep{silberer2014learning},
\textit{VisSim}~\citep{silberer2014learning}.
\textit{MEN} was  designed to capture general word  ``relatedness.'' \textit{Simlex-999} and \textit{SemSim} measure notions of semantic similarity, and \textit{VisSim} ranks the same words as \textit{SemSim} but in terms of visual similarity.
In each case, the designers of the benchmarks provided pairs of words to human judges, who in turned provide ratings based on the metric of the benchmark.
To judge our model, we calculate the cosine similarity of our embeddings for the word pairs and then calculate Spearman's \(\rho\) between our list of ratings and those of the human judges.
We evaluate three versions of our model on these benchmarks: pseudowords using the centroid method (\textsc{Psuedowords-C}), pseudowords using the hypersphere method (\textsc{Pseudowords-H}), and the centroid method with a randomly initialized mapping (\textsc{Pseudowords-Ran}), as explained below.

\item[Existing Multimodal Models] We compare our results on these benchmarks against previously published results for other multimodal word embeddings . Using the results published by~\citet{lazaridou2015combining} and a target word embedding of 300, we compare our results to their \textsc{MMSkip-gram-A} and \textsc{MMSkip-gram-B}, which maximize the similarity of the textual and visual representations under a max-margin framework.; the former constrains the dimensionality of the visual features to be the same as the word embeddings, while the latter learns an explicit mapping between the textual and visual spaces.
We also include baseline results for pure-text skip-gram embeddings (\textsc{Skip-gram})).
\end{description}

\begin{table*}[t]
  \centering
  \begin{tabular}{l|c|c|c|c}
    \textbf{Model} & \textit{MEN} & \textit{Simlex-999} &
                                                          \textit{SemSim}
    & \textit{VisSim} \\ \hline 
    \textsc{MMSkip-gram-A} & \textbf{0.75} & 0.37 & \textbf{0.72} & \textbf{0.63} \\ 
    \textsc{MMSkip-gram-B} & 0.74 & \textbf{0.40} & 0.66 & 0.60 \\ \hline
    \textsc{Skip-gram} & 0.70 & 0.33 & 0.62 & 0.48 \\ \hline
    \textsc{Pseudowords-Ran} & 0.51 & 0.28 & 0.34 & 0.26 \\ \hline
    \textsc{Pseudowords-C} & 0.70 & 0.34 & 0.70 & 0.53 \\
    \textsc{Pseudowords-H} & 0.70 & 0.34 & 0.71 & 0.53 \\ \hline
  \end{tabular}
  \caption{Spearman correlation between the generated multimodal
    similarities and the benchmark human judgments. In all cases,
    results are reported on the full set of word similarity pairs.}
  \label{tab:spearman}
\end{table*}

\subsection{Results}
\label{sec:results}
The results for the human judgment experiments are presented in Table~\ref{tab:spearman}.
For these experiments, we tried two methods of initializing the mapping.
First, \textit{Random Initialization:} the visual-textual mapping matrix was randomly initialized in the same manner as the word embeddings, with the goal of allowing the mapping to be freely generated from the word context.
Second, \textit{Neural Weight Initialization:} to boost the performance of the multimodal embeddings, the mapping was initialized with the weights from a simple neural network trained to predict known word embeddings \footnote{We used embeddings from the Google News dataset available at https://code.google.com/p/word2vec.} from our convolutional image features.

\subsubsection{Random Initialization}

Interestingly, there is a degradation in the correlation from the
addition of the visual features across all benchmarks.
This seems to indicate that the induced mapping, when beginning with a random initialization, is yet
insufficient to properly situate the convolutional features into the
multimodal space.
It would seem initially that, during training, while the mapping is
still being learned, adding the visual context to the text vectors
perhaps worsens the representational quality of the word embedding.

\subsubsection{Neural Weight Initialization}

On the other hand, when the mapping is quickly pretrained on existing distributed word representations, the results are greatly improved.
In the cases of capturing general relatedness and pure visual similarity, the multimodal model of~\citet{lazaridou2015combining} performs better.
However, in the case of capturing semantic word similarity, our model performs signficantly better than \textsc{MMSkip-gram-B} (although it should be noted that these results are roughly on par with the benchmark authors~\citep{silberer2014learning} and a point below the non-mapping \textsc{MMSkip-gram-A}).
Although further work is needed to examine this result, the performance of the model in this case can be visualized through an example.
Table~\ref{tab:words} provides some insights on the changes made to the word embeddings as a result of the inclusion of visual information in the learning process. In the two visual instances, our model captures many of the same nuances as \textsc{MMSkip-gram-B} over the \textsc{Skip-gram} model: \textit{donuts} are more similar to other types of food than to places where you find donuts and \textit{owls} are more similar to other birds of prey than just woodland creatures. However, our model seems to capture more of the semantic idea of donuts as ``junk food'' rather than just the visual similarity of roundness (the link established between donut and cupcake is particularly interesting). As for \textit{owl}, some of the visual similarity is lost, by ranking \textit{sparrow} first, with regards to the class of birds of prey, but there seems to be a recognition of the semantic relationship between the top similar words (``sparrow hawk'' is a synonym for ``kestrel'' in Wordnet, for example) as well as visual similarity via brown feathers and beaks.

As for the representations learned without explicit visual information, our model still seems to demonstrate the propagation of this information but in a different manner than \textsc{MMSkip-gram-B}. The words ranked as similar to \textit{mural} lose the artistic concepts of painting and portrait ranked highly by the other models; instead our model ranks ``fresco'' and ``bas-relief'' alongside sculpture, capturing instead a more complex representation of ``artwork executed directly on a wall.'' For \textit{tobacco}, our model dismisses the recreational uses of tobacco captured via ``cigar'' and ``cigarette,'' while also ignoring the na\"ive ``crop'' sense captured by ``corn.'' Instead, the highly-ranked words seem to display a more robust use of tobacco in the semanic sense as a \textit{cash crop}, specifically referencing other notable trade crops.\footnote{Hemp and coffee are also ranked as more similar than cigarettes.}

The two abstract concepts, \textit{depth} and \textit{chaos}, reveal two very different results of the visual propagation. For \textit{depth}, no evidence of the visual is apparent: unlike the relation to the sea drawn by \textsc{MMSkip-gram-B}, our model seems to only capture \textit{depth}'s semantic similarity to other types of measurement (height and thickness). For \textit{chaos}, there is still the loss of more imageable concepts, as in shadow, from the rankings; however, our model instead seems to capture words more semantically similar to \textit{chaos} (anarchy) or synonyms of events like \textit{chaos} (turmoil and pandemonium).

\begin{table*}[t]
\centering
\begin{tabular}{l|c|c|c}
\textit{Word} & \textsc{Skip-gram} & \textsc{MMSkip-gram-B} & \textsc{Pseudowords (H)} \\ \hline
donut & fridge, diner, candy & pizza, sushi, sandwich & sandwich, candy, cupcake\\
owl & pheasant, woodpecker, squirrel & eagle, falcon, hawk & sparrow, kestrel, hawk\\ \hline
mural & sculpture, painting, portrait & painting, portrait, sculpture & fresco, bas-relief, sculpture\\
tobacco & coffee, cigarette, corn& cigarette, cigar, smoking& cotton, cocoa, sugar\\
depth & size, bottom, meter& sea, size, underwater& height, surface, thickness\\
chaos &anarchy, despair, demon & demon, anarchy, shadow& anarchy, turmoil, pandemonium
\end{tabular}
\caption{Top 3 neighbors of the target words, ordered by similarity. The training data contained visual information only for \emph{donut} and \emph{owl}}
\label{tab:words}
\end{table*}

\section{Conclusion}

Our model performs multimodal skip-gram using pseudowords construction from convolutional image features, and then demonstrates the propagation of visual information to non-visual words but in a seemingly distinct manner from the existing models.
Of particular note is that it is apparent that distributed word representations are being improved and informed by this information over pure-text skipgram, but these embeddings seem to perform best at a semantic rather than a visual level.
Future work will focus on the nature of this embedding.
In particular, we will investigate the applicability of the induced textual-visual mapping to the task of zero-shot image labeling~\citep{socher2013zero} and image retrieval.
As of yet, the nature of the mapping, beyond the qualitative improvements provided to the word embeddings, is still unclear, and future work will seek to address this.

\bibliography{main}
\bibliographystyle{iclr2016_conference}

\end{document}